\documentclass{article}

    \PassOptionsToPackage{numbers, compress}{natbib}

\usepackage[preprint]{neurips_2024}

\usepackage{microtype}
\usepackage{graphicx}
\usepackage{subfigure}
\usepackage{booktabs} 
\usepackage{tabularx}
\usepackage{hyperref}
\usepackage{pifont}

\usepackage[utf8]{inputenc} 
\usepackage[T1]{fontenc}    
\usepackage{hyperref}       
\usepackage{url}            
\usepackage{booktabs}       
\usepackage{amsfonts}       
\usepackage{nicefrac}       
\usepackage{microtype}      
\usepackage{xcolor}         

\usepackage{times}
\usepackage{epsfig}
\usepackage{graphicx}
\usepackage{amsmath}
\usepackage{amssymb}

\usepackage{capt-of}

\usepackage{amsmath}
\usepackage{amssymb}
\usepackage{mathtools}
\usepackage{amsthm}

\usepackage{placeins} 
\usepackage{afterpage}
\usepackage{stfloats}
\usepackage{enumitem}
\usepackage[most]{tcolorbox} 

\theoremstyle{plain}

\theoremstyle{definition}

\theoremstyle{remark}

\newcommand{\dcircle}[2][0.4]{%
  \tikz[baseline=(char.base)]{
    \node[shape=circle, draw=black, fill=black,
          minimum size=2*#1 cm, inner sep=0pt] (char)
          {\textcolor{white}{#2}};
  }%
}
\newcommand{\agentlogo}{\raisebox{-0.5em}{\includegraphics[height=2.0em]{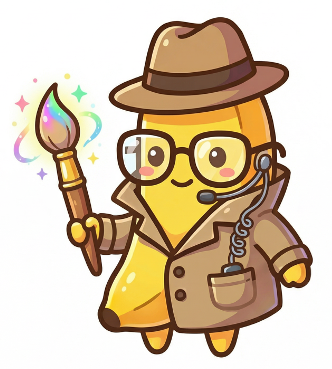}}}
\definecolor{abMid}{RGB}{255,140,0}      
\definecolor{abEnd}{RGB}{100,156,230} 

\usepackage{xcolor}
\usepackage{xspace}

\definecolor{MagicPurple}{RGB}{142, 68, 173}  
\definecolor{MagicPink}{RGB}{233, 30, 99}    

\definecolor{BananaOrange}{RGB}{255, 152, 0}  
\definecolor{BananaYellow}{RGB}{255, 235, 59} 

\def\namecolor{%
  \textbf{%
    \textcolor{MagicPurple}{A}%
    \textcolor{MagicPurple!70!MagicPink}{g}%
    \textcolor{MagicPurple!40!MagicPink}{e}%
    \textcolor{MagicPink!80!MagicPurple}{n}%
    \textcolor{MagicPink}{t}%
    \kern.2em%
    \textcolor{BananaOrange}{B}%
    \textcolor{BananaOrange!90!BananaYellow}{a}%
    \textcolor{BananaOrange!75!BananaYellow}{n}%
    \textcolor{BananaOrange!60!BananaYellow}{a}%
    \textcolor{BananaOrange!40!BananaYellow}{n}%
    \textcolor{BananaOrange!20!BananaYellow}{a}%
  }\xspace%
}

\title{\agentlogo \namecolor: High-Fidelity Image Editing with Agentic Thinking and Tooling}

\author{%
Ruijie Ye$^{1,2\dagger}$, Jiayi Zhang$^{3\dagger}$, Zhuoxin Liu$^{4\dagger}$, Zihao Zhu$^{1}$, Siyuan Yang$^1$,\\ 
\bf Li Li$^5$, Tianfu Fu$^6$, Franck Dernoncourt$^{7\ddagger}$, Yue Zhao$^5$, Jiacheng Zhu$^{8\mathsection}$,\\
\bf Ryan Rossi$^{7\ddagger}$, Wenhao Chai$^9$, Zhengzhong Tu$^{1\star}$
\\[2pt]
$^1$TAMU\quad $^2$Brown University\quad $^3$UW-Madison\quad $^4$UCSD\quad
$^5$USC\quad
$^6$xAI\\
\quad$^7$Adobe Research\quad $^8$Meta AI
\quad$^9$Princeton University
\\[2pt]
\small $^\star$Corresponding Author: \texttt{tzz@tamu.edu}. $^{\dagger}$Equal contributions.\\[2pt]
$^{\ddagger}$Work not done at Adobe. 
$^{\mathsection}$Work done outside of Meta.\\[2pt]
\textbf{\textcolor{magenta}{Project Website}}: \href{https://agent-banana.github.io}{\color{black}{\texttt{agent-banana.github.io}}}
}

\begin{document}
\maketitle

\vspace{-5mm}
\noindent\makebox[\textwidth][c]{%
    \begin{minipage}{1.1\textwidth}
        \includegraphics[width=\textwidth]{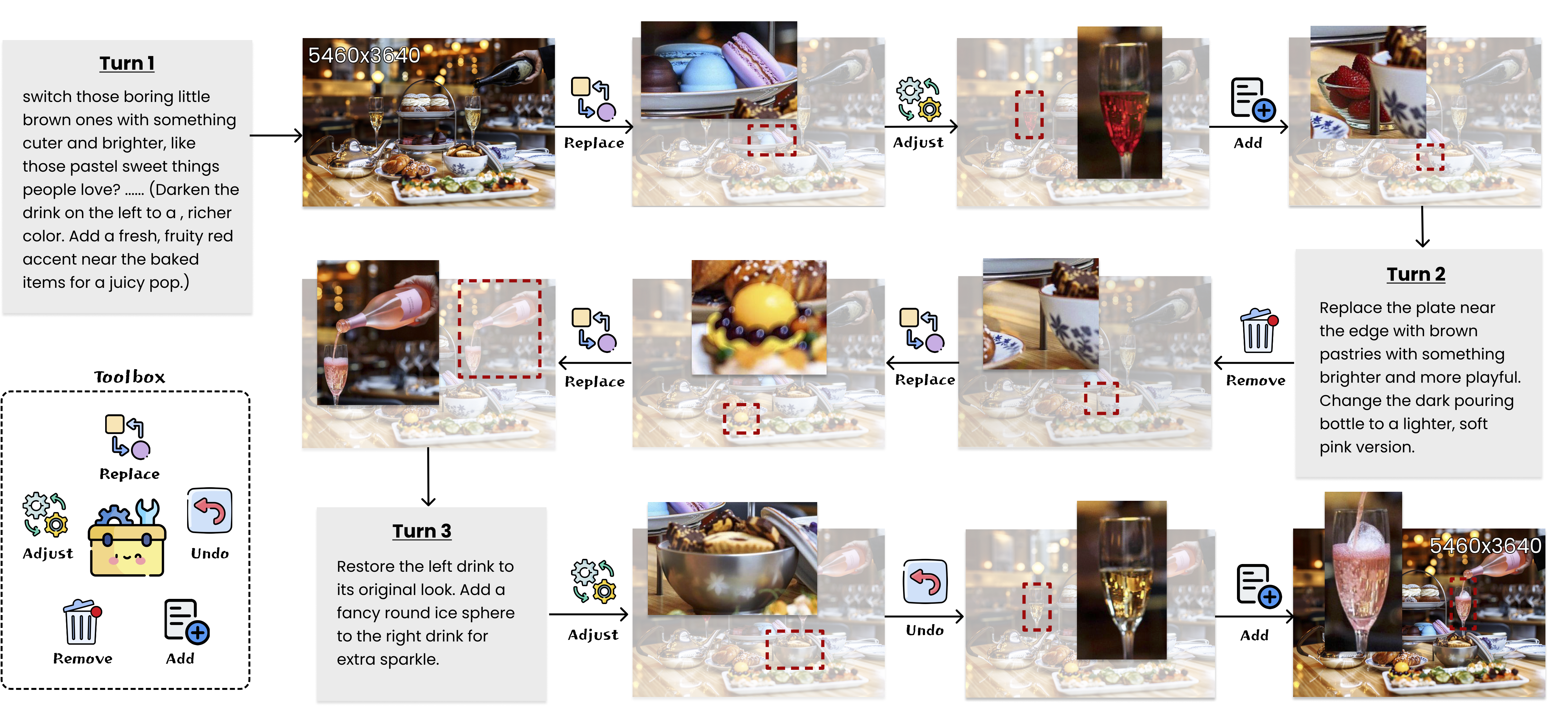}
        \vspace{-5mm}
        \captionof{figure}{We present
        \textbf{\namecolor}, an agentic editing system that enables \textbf{high-fidelity, native-resolution} image editing through \textbf{reasoning-based} natural-language interaction, where each edit is \textbf{context-aware}, \textbf{logically dependent}, and \textbf{locally precise}.
In this example, the user provides a vague yet complex editing prompt, and Agent Banana iteratively refines a scene in native high resolution ($5460\times 3640$)—from a \textbf{stylistic replacement} (\textbf{Turn 1}), to \textbf{attribute decoupling} that preserves non-target dynamics (changing the bottle color without affecting the pouring liquid; \textbf{Turn 2}), and finally to \textbf{retrieving prior state} and adding fine details (\textbf{Turn 3}).
The result is a professional-style workflow that resists \textbf{over-editing} and \textbf{background drift}, while faithfully preserving what should remain unchanged.}
        \label{fig:teaser}
    \end{minipage}
}

\begin{abstract}
We study instruction-based image editing under professional workflows and identify three persistent challenges: \textbf{{(i)}} editors often over-edit, modifying content beyond the user’s intent; \textbf{(ii)} existing models are largely single-turn, while multi-turn edits can alter object faithfulness; and \textbf{(iii)} evaluation at around 1K resolution is misaligned with real workflows that often operate on ultra high-definition images (e.g., 4K). 
We propose \textbf{\namecolor}, a hierarchical agentic planner–executor framework for high-fidelity, object-aware, thinking with editing. 
\textbf{\namecolor} introduces two key mechanisms: \textbf{\ding{182} Context Folding}, which compresses long interaction histories into structured memory for stable long-horizon control, and \textbf{\ding{183} Image Layer Decomposition}, which performs localized layer-based edits to preserve non-target regions while enabling native-resolution outputs. To support rigorous evaluation, we build HDD-Bench, a high-definition, dialogue-based benchmark featuring verifiable stepwise targets and native 4K images (11.8M pixels) for diagnosing long-horizon failures. On HDD-Bench, Agent Banana achieves the best multi-turn consistency and background fidelity (e.g., IC 0.871, SSIM$_\text{OM}$ 0.84, LPIPS$_\text{OM}$ 0.12) while remaining competitive on instruction following, and also attains strong performance on standard single-turn editing benchmarks. We hope this work advances reliable, professional-grade agentic image editing and its integration into real workflows.
\end{abstract}


\section{Introduction}

Instruction-based image editing~\cite{brooks2023instructpix2pix,zhang2023internlm,wang2025seededit,deng2025emerging,labs2025flux,cai2025hidream,openai2025imagegen,wu2025omnigen2,liu2025step1x} enables users to modify images via natural-language commands and has become a core capability of modern generative vision systems. Recent advances in foundation models---particularly diffusion~\cite{ho2020denoising,liu2022compositional} and autoregressive transformers~\cite{wang2024genartist}---have substantially improved both photorealism and instruction following, powering practical editing experiences in commercial systems (e.g., GPT-4o~\cite{openai2025imagegen}, Gemini 2.5 Flash Image~\cite{google2025gemini25flashimage}) and strong open-source models (e.g., Flux-1~\cite{labs2025flux1kontextflowmatching}, Qwen-Image-Edit~\cite{wu2025qwen}).

Despite this rapid progress, a substantial gap remains between current generative editors~\cite{wu2025qwen,liu2025step1x,hui2024hqedit} and the requirements of \emph{professional} workflows. In high-stakes settings such as photography~\cite{hu2025real}, graphic design~\cite{mirzaei2025impact}, visual effects (VFX), and filmmaking~\cite{zhang2025generative}, users typically work on native high-resolution assets (often 4K or higher) and demand precise, localized modifications that preserve all non-target content~\cite{hui2024hqedit}. By contrast, today’s models often operate at reduced resolution or rely on downsampling, making it difficult to maintain fine textures and sharp boundaries.
Moreover, they frequently exhibit over-editing effects, unintentionally altering regions outside the user’s intent or degrading global semantic coherence.
Lastly, they struggle with complex requests that are multi-goal or sequential~\cite{zhou2025multi}, where success requires decomposing the instruction, verifying intermediate results, and revising earlier decisions across turns.

We argue that to bridge this gap, next-generation editing tools must satisfy four core capabilities: \dcircle[0.15]{1} \textbf{Intent understanding} and decomposition of complex requests into atomic sub-edits; \dcircle[0.15]{2} \textbf{Accurate localized editing} to ensure edits are precisely applied while maintaining the rest of the content unchanged, on native resolution; \dcircle[0.15]{3} \textbf{State tracking and rollback} to retain intermediate steps across multi-turn interactions so that users (or intelligent agents) can easily revert to a previous step and re-plan the remaining steps; and  \dcircle[0.15]{4} \textbf{High-resolution native editing} to operate directly on native 4K images, preserving fine-grained textures and sharp boundaries while avoiding downsampling.

To this end, we introduce \textbf{\namecolor}, an agentic, layer-aware image editing framework that couples high-level reasoning and planning with tool-use capabilities, benefiting from the rapid progress of Vision-Language Models (VLMs) in image understanding, reasoning, and tool invocation~\cite{hong2024cogagent,qin2025ui-tars, wu2024atlas, xu2024aguvis,openai2025operator,anthropic2025claude37sonnet}.
Agent Banana decomposes ``vibe''-type prompts into discrete, single-goal steps, executes these steps using a `Photoshop-style` layer isolation, masking, and local edits. 
Agent Banana also includes a self-reflection mechanism~\cite{yao2022react, shinn2023reflexion}, allowing it to retry, rollback, and replan at inference time.
Crucially, Agent Banana is built around two mechanisms tailored for long-horizon, high-resolution editing: Context Folding, which compresses long interaction histories into structured memory for stable state tracking across turns, and Image Layer Decomposition, which performs edits on isolated high-resolution layers to preserve non-target content and prevent drift across iterations.

To evaluate multi-turn, high-definition editing under realistic stepwise dependencies, we build HDD-Bench, a High-Definition and Dialogue-based benchmark designed to simulate professional editing workflows. Unlike prior benchmarks that are predominantly single-turn or weakly dependent across turns~\cite{deng2025emerging,labs2025flux,cai2025hidream,wu2025omnigen2,liu2025step1x,ye2025imgedit}, HDD-Bench features logically dependent instruction chains where each turn induces a well-defined state transition and can be verified step by step. HDD-Bench benchmarks instruction adherence, edit locality, multi-turn consistency, and overall visual fidelity at native resolution. To reduce evaluation ambiguity, we further introduce a graph-based evaluation protocol that tracks object-state transitions across turns, complementing global perceptual metrics with localized, turn-level checks of whether the intended edits are applied and whether non-target regions remain preserved.

\begin{figure*}[t]
  \centering
  \hspace*{-0.01\linewidth}\includegraphics[width=0.98\linewidth]{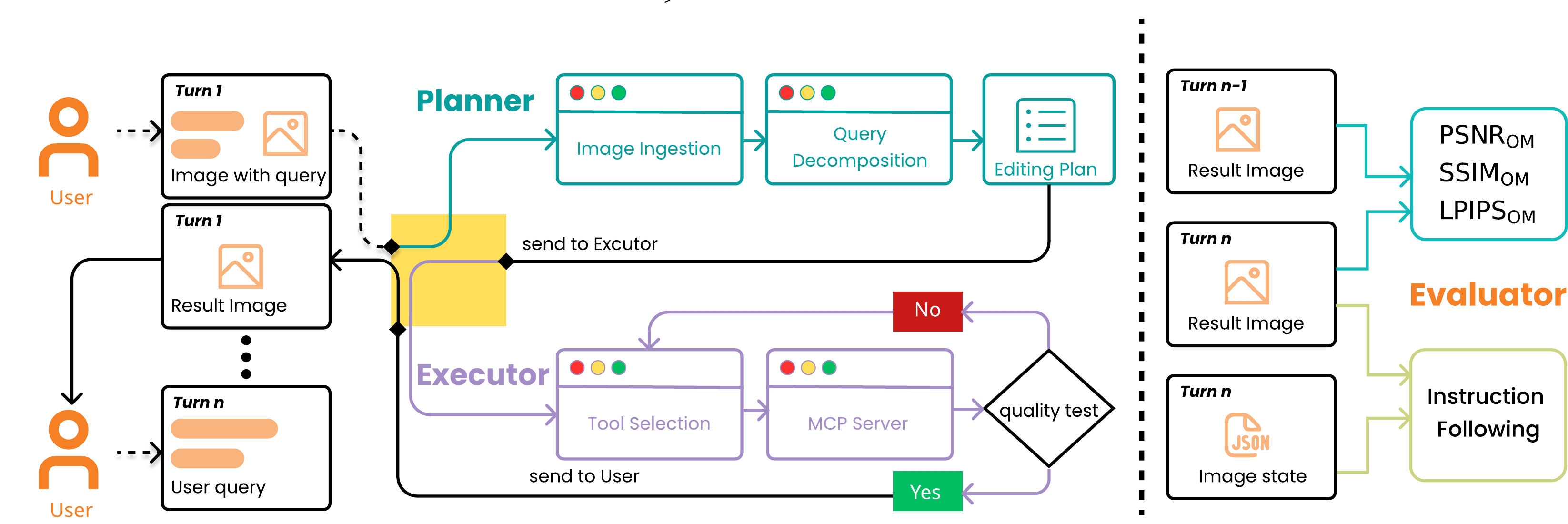}
  \vspace{-2mm}
  \caption{\textbf{Overview of the Agent Banana Framework.} 
  The system operates in a multi-turn loop (Left), comprising two core agents: a \textbf{Planner} that decomposes user queries into executable editing plans, and an \textbf{Executor} that selects tools via the MCP Server. 
  Crucially, the Executor incorporates a \textbf{self-correction mechanism} (Quality Test), reiterating the editing process if the quality check fails before presenting the result to the user.
  (Right) Our \textbf{Evaluator} assesses performance by analyzing the transition between Turn $n-1$ and Turn $n$, utilizing instruction adherence checks and state tracking (JSON) to derive the final score.}
  \label{fig:workflow}
\end{figure*}

\section{Agent Banana Framework}

\subsection{Problem Setup and Motivation}

We consider a multi-turn instruction-based image editing task, where the user provides a sequence of natural-language instructions $q = \{q_1, q_2, \dots, q_T\}$ and an initial image $I_0$. The system responds by executing a trajectory of editing steps $\tau = \{(a_1, o_1), (a_2, o_2), \dots, (a_T, o_T)\}$, where each $a_i$ denotes the $i$-th action (comprising reasoning and tool invocation), and $o_i$ is the resulting image state. The environment dynamics can be abstracted as a transition operator $\mathcal{E}$ such that $o_i = \mathcal{E}(o_{i-1}, a_i)$. Following the ReAct-style paradigm~\cite{yao2022react}, the agent incrementally selects actions based on the full interaction history:
\begin{equation}
    P_{\theta}(\tau \mid q) \propto \prod_{i=1}^{T} \pi_{\theta}\left(a_i \mid q, a_{<i}, o_{<i}\right).
\end{equation}

While conceptually simple, this formulation introduces two major challenges in practice:
\ding{182} \textbf{Long-horizon context overflow}. As the number of editing steps increases, the agent must repeatedly condition on the entire interaction history, both textual and visual. This leads to severe token inefficiency, quickly exceeding the LLM’s context length, and introduces irrelevant noise that impairs reasoning and planning in later steps.
\ding{183} \textbf{Full-image detail degradation} Existing editing tools often operate by resampling the entire image at each step, regardless of the locality of the edit. This not only wastes computation on unchanged regions, but also causes subtle degradation of fine details over time—especially in backgrounds or fixed objects—leading to accumulation of visual artifacts across turns.




\subsection{Overview of Agent Banana}

To address the challenges of context overflow and iterative degradation, we introduce \textbf{Agent Banana}, a hierarchical multi-agent editing framework designed for high-fidelity, multi-turn image editing at native resolution, as shown in
Figure~\ref{fig:workflow}. The framework explicitly separates global task reasoning from low-level execution via two specialized agents:
\begin{itemize}[leftmargin=*,topsep=0pt]
    \setlength{\itemsep}{0pt}
    \setlength{\parskip}{0pt}
    \setlength{\parsep}{0pt}
    \setlength{\itemsep}{0pt}
    \vspace{-2mm}
    \item \textbf{Planner}: Performs global intent interpretation, decomposes complex user instructions into executable sub-goals, and monitors overall progress.
    \item \textbf{Executor}: Carries out atomic editing operations, invokes tools on localized image regions, and handles intermediate validation and error recovery.
    \vspace{-2mm}
\end{itemize}

This division of roles enables the system to both reason over long-horizon objectives and execute fine-grained visual edits in a scalable and interpretable manner.

Agent Banana is built around two key mechanisms that mitigate the core bottlenecks identified earlier:
\begin{itemize}[leftmargin=*]
    \setlength{\itemsep}{0pt}
    \setlength{\parskip}{0pt}
    \setlength{\parsep}{0pt}
    \setlength{\itemsep}{0pt}
    \vspace{-2mm}
    \item \textbf{Context Folding:} A hierarchical memory abstraction that compresses the growing interaction history into structured representations, enabling long-horizon planning without exceeding context limits.
    \item \textbf{Image Layer Decomposition (ILD):} A localized execution strategy that performs edits on cropped high-resolution patches (layers), preserving pixel-level fidelity in unedited regions and naturally supporting ultra-HD editing workflows.
    \vspace{-2mm}
\end{itemize}

During each interaction round, the Planner receives the user instruction and current image state, decomposes the task into sub-goals, and delegates them to the Executor. The Executor generates intermediate candidates via ILD-based editing and returns feedback. The Planner verifies whether the updated image meets the instruction goal; if not, it can replan or rollback using the maintained image state graph. This closed-loop process continues until the user's intent is satisfied or a predefined turn limit is reached.

\subsection{Context Folding}

To effectively mitigate the exponential explosion of context in long-horizon tasks, we introduce the \textbf{Context Folding} mechanism. The core idea is to "fold" the raw, high-dimensional interaction history into a compact semantic representation through hierarchical abstraction and selective memory. Specifically, we decouple context information into three schemas of varying granularity: the Asset Level, the Execution Level, and the Planning Level.

\paragraph{Asset Level: ImageContext.}
This is the fundamental data unit of the system, constructed by the Executor after each image generation. Instead of directly embedding high-dimensional image tokens, ImageContext abstracts the image into a lightweight semantic node, containing a unique identifier (URI), VLM-generated textual description of the content, its parent URI, and the transformation type leading to this state change. Through this text-based graph representation, the agent can track the full image evolution history with minimal context overhead while preserving the topological relationships between image states.

\paragraph{Execution Level: ToolContext.}
This serves as the \textbf{Transient Working Memory} used by the Executor during single-step reasoning. It details the microscopic operations required to complete an atomic instruction, including tool selection, parameter configuration, the intermediate reasoning process (Thought), and references to relevant ImageContexts. ToolContext primarily facilitates error recovery and state backtracking within the current step. Once the current sub-task is completed, these trivial trial-and-error details are "folded" and do not enter the long-term global memory, thereby preventing irrelevant execution noise from interfering with the Planner.

\paragraph{Planning Level: ActionContext.}
This forms the \textbf{Persistent Memory} established after each round of user interaction. When the Planner confirms that a series of operations successfully meets the user's requirements, it constructs an ActionContext. This context retains only the verified effective editing path: the final intention determined by the Planner and the corresponding sequence of key ImageContexts. ActionContext essentially acts as a semantic compression of ToolContext, discarding procedural tool invocation details and preserving only high-level task semantics and result states. This ensures that the agent maintains a clear cognitive grasp of the global task state even after dozens of interaction turns, without being overwhelmed by excessive token sequences.

\subsection{Image Layer Decomposition}

To resolve the issues of detail loss and resolution limitations inherent in full-image generation, we propose the \textbf{Image Layer Decomposition (ILD)} mechanism. Traditional end-to-end editing models often resample the entire image, causing unintended pixel drift in unedited regions (such as the background or irrelevant objects). The ILD mechanism abandons this global operation in favor of a "decompose-edit-fuse" local processing paradigm.

Specifically, this mechanism utilizes a dynamic object-aware mask to precisely localize the target region, losslessly cropping it from the original high-resolution image into an independent layer patch. All generative editing is performed solely within the local coordinate system of this patch, thereby freezing the pixel state of the background region and substantially reducing degradation in non-target regions by avoiding full-image resampling. Upon completion of editing, the system seamlessly blends the updated patch back into the original image using Gaussian blending algorithms. Furthermore, since it only processes local patches, this mechanism naturally supports ultra-high-definition image editing beyond the model's native resolution limits.

Based on the ILD mechanism, we define an \textbf{Action Space} of five atomic instructions that cover common editing needs:

\begin{itemize}[leftmargin=*]
    \setlength{\itemsep}{0pt}
    \setlength{\parskip}{0pt}
    \setlength{\parsep}{0pt}
    \vspace{-2mm}
    \item \textbf{\texttt{replace}}: Substitutes the content of the target layer with a new object using inpainting techniques while maintaining edge consistency.
    \item \textbf{\texttt{remove}}: Eliminates the target layer and fills the void using background completion algorithms.
    \item \textbf{\texttt{add}}: Generates a new layer at a specified location and performs layer superposition.
    \item \textbf{\texttt{adjust}}: Applies attribute transformations (e.g., color correction, style transfer) to the target layer without altering its geometry.
    \item \textbf{\texttt{undo}}: Rapidly rolls back to the previous image state node based on the state graph maintained in Context Folding.
\end{itemize}
\vspace{-2mm}

These five atomic operations form the foundational capability set for Agent Banana, enabling the Planner to execute complex, composite edits by composing these primitives.

\begin{table}[t]
    \centering
        \caption{\textbf{Comparison of existing image editing datasets vs.\ our HDD-Bench.} We compare key features including support for multi-turn interaction, high-resolution images, object-level editing granularity, reasoning capabilities, and ground-truth verification. HDD-Bench is the only benchmark encompassing all these capabilities, bridging the gap for professional-grade editing evaluation.}
    \label{tab:rw_compare}
      \begin{tabular}{@{}lccccc@{}}
        \toprule
        Dataset & \#Turn & Res. (pxs) & Obj Scale & Reason. & Verifi. \\
        \midrule
        AnyEdit~\cite{yu2025anyedit}       & 1.0 & 0.4M  & Large  & --         & -- \\
        GEdit-Bench~\cite{liu2025step1x}   & 1.0 & 3.4M  & Large  & --         & -- \\
        SEED-DataEdit~\cite{ge2024seed}    & 3.8 & 1.1M  & Medium & \checkmark & -- \\
        ImgEdit-Bench~\cite{ye2025imgedit} & 1.0 & 1.1M  & Large  & --         & -- \\
        \hline
        \textbf{HDD-Bench (ours)}          & 3.0 & 11.8M & Small  & \checkmark & \checkmark \\
        \hline
      \end{tabular}

\end{table}

\section{HDD-Bench: High-Definition, Dialogue-based image editing benchmark}

Recent generative editors are increasingly interactive and agentic, yet rigorous evaluation for \emph{professional-grade} editing remains underdeveloped. Existing benchmarks typically fall short along at least one key dimension: (i) \emph{single-turn interactions} that fail to capture the stepwise dependencies inherent to real editing sessions; (ii) \emph{low-resolution formats} that cannot meet the fidelity and locality requirements of native 4K workflows; and (iii) \emph{human-in-the-loop processes} that, while enabling richer interactions, act as a significant bottleneck restricting dataset scale and diversity. 
More importantly, most benchmarks provide only an end result, without a \emph{verifiable intermediate interface}. Without turn-by-turn targets, it is hard to diagnose long-horizon failures such as error accumulation, over-editing of non-target regions, or semantic drift across turns. This motivates a benchmark that (i) supports \emph{multi-turn}, logically dependent instruction chains; (ii) evaluates at \emph{native high resolution} to ensure fidelity; and (iii) provides \emph{structured intermediate supervision}, enabling precise failure attribution without the scalability constraints of human oversight.

\subsection{A Scalable Data Pipeline for Multi-turn Editing}

\begin{figure}[t]
    \centering
    \includegraphics[width=\linewidth]{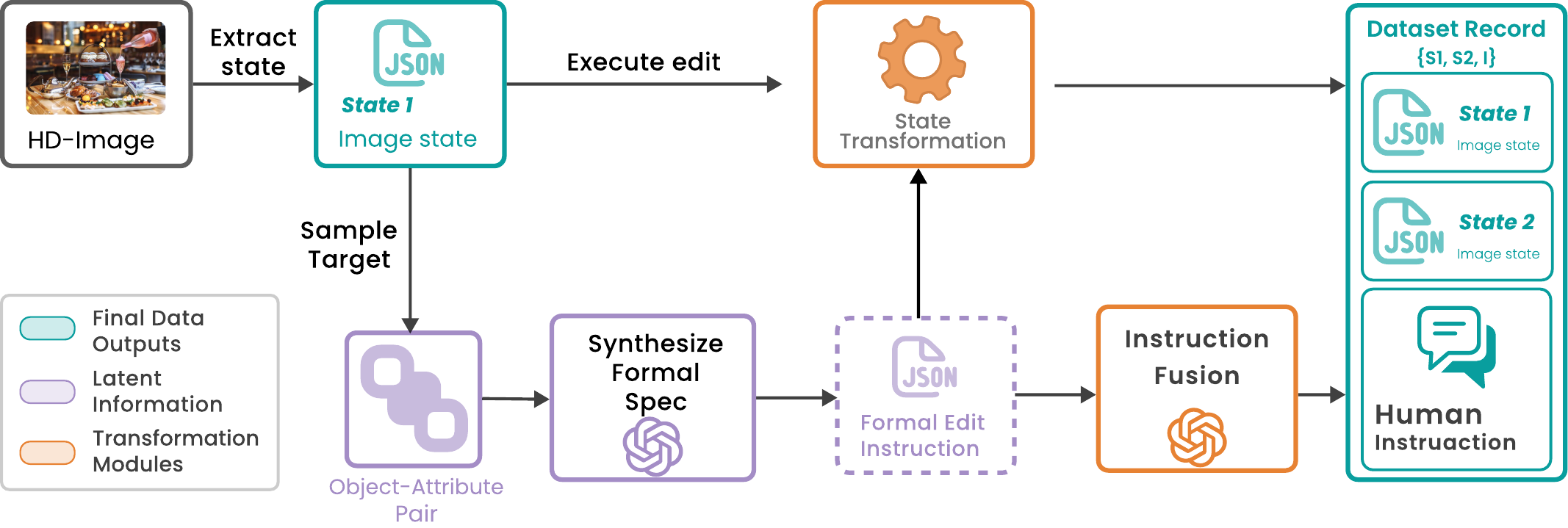}
        \vspace{-6mm}
    \caption{\textbf{Scalable Data Pipeline for Multi-turn Editing.}
    This diagram illustrates the process of generating aligned (State, Instruction) pairs from HD images.
    }
    \label{fig:engine}
  \end{figure}

To enable verifiable multi-turn supervision without expensive pixel-level annotation, we propose a \textbf{scalable symbolic data engine} that synthesizes editing trajectories in an attribute-level state space. For each input image, we construct an initial scene state $s_0$ that represents salient objects and their attributes, including \texttt{name}, \texttt{color}, \texttt{size}, \texttt{material}, and \texttt{shape}. Each editing turn is specified by a set of \emph{canonical} edit commands $\mathbf{c}_t$, which are applied deterministically to update the state:
\[
s_{t+1} = \mathcal{T}(s_t, \mathbf{c}_t),
\]
where $\mathcal{T}$ is a deterministic transition operator that modifies only the targeted object attributes. This design decouples \emph{interaction synthesis} from \emph{image generation}: we can generate consistent and checkable intermediate targets $\{s_1, s_2, \dots\}$ without rendering images during data construction.

To mimic real user behavior, a language agent paraphrases the canonical command set $\mathbf{c}_t$ into a single natural-language instruction $q_t$, optionally mixing multiple intents (e.g., adding an object while changing another object’s color). Importantly, any ambiguity is introduced only in the surface phrasing $q_t$, while the underlying $\mathbf{c}_t$ and target state $s_{t+1}$ are preserved as the internal ground truth. 
To ensure reliability, we incorporate human verification at the entry point of the pipeline: the initial scene graph and extracted attributes used to form $s_0$ are manually inspected and corrected. Since subsequent turns are produced by deterministic transitions, this guarantees the correctness of the entire multi-turn state chain and provides a principled, verifiable interface for evaluation.

\subsection{Constructing the HDD-Bench}

Built on top of the data engine, we construct \textbf{HDD-Bench}, a High-Definition, Dialogue-based benchmark that targets professional editing requirements. HDD-Bench is designed to jointly stress (i) \emph{multi-turn dependency}, where later instructions build on earlier edits; (ii) \emph{high-resolution fidelity}, where fine textures and sharp boundaries must be preserved at native resolution; and (iii) \emph{object-level compositionality}, where instructions may involve multiple objects and mixed intents.

Each sample in HDD-Bench is a \textbf{three-turn} editing session. At turn $t$, the benchmark provides a natural-language instruction $q_t$ (often combining multiple edit intents into a single request) and a corresponding target symbolic state $s_t$ for verifiable evaluation. We adopt three-turn interactions to control difficulty and simplify comparisons across methods, while still capturing stepwise dependency and error accumulation; notably, our engine can generate longer sessions without changing the evaluation interface.

HDD-Bench contains \textbf{96} curated sessions selected from the synthesis pipeline. The selected samples emphasize scenes with multiple salient objects and non-trivial edit chains, and cover a diverse set of atomic actions (e.g., \texttt{add}, \texttt{remove}, \texttt{replace}, \texttt{adjust}, \texttt{undo}) as well as hybrid instructions that require composing multiple actions within a turn.

\subsection{Evaluation Protocol}

HDD-Bench evaluates editing quality from two complementary perspectives: (i) \textbf{semantic correctness} of the intended edits, and (ii) \textbf{visual preservation} of non-target regions. The first aspect is assessed in a verifiable, object-centric manner using the symbolic state representation; the second is assessed at the pixel/perceptual level to quantify background fidelity.

\paragraph{State-based metrics: Instruction Following and Image Consistency.}
Given a generated image at turn $t$, we map it to a predicted post-edit state $\hat{s}_t$ using the same perception pipeline used to construct $s_0$. We then compare $\hat{s}_t$ against the ground-truth target state $s_t$ to compute two scores:
\textbf{Instruction Following (IF)} measures whether the attributes of \emph{targeted} objects match the requested edits, while
\textbf{Image Consistency (IC)} measures whether \emph{non-target} objects remain unchanged across turns.
Both scores are computed by averaging attribute-level correctness over objects:
\[
s_{\text{IF}}\ \text{or}\ s_{\text{IC}}
= \frac{1}{N} \sum_{i=1}^{N}
\left(
\frac{1}{M_i} \sum_{j=1}^{M_i} s_{i,j}
\right),
\]
where $N$ is the number of evaluated objects (edited or preserved), $M_i$ is the number of attributes for object $i$, and $s_{i,j}$ is the correctness score for the $j$-th attribute.

\paragraph{Otsu-masked background fidelity.}
Global full-reference metrics such as PSNR, SSIM, and LPIPS can be misleading for editing, since they penalize valid foreground changes and unwanted background corruption equally. To isolate preservation quality, we compute \textbf{Otsu-Masked PSNR/SSIM/LPIPS}~\cite{otsu1975threshold,wang2004image,zhang2018perceptual}, denoted as $\mathrm{PSNR}_{\mathrm{OM}}$, $\mathrm{SSIM}_{\mathrm{OM}}$, and $\mathrm{LPIPS}_{\mathrm{OM}}$. Concretely, we form a pixel-wise difference map between the pre-edit and post-edit images, apply Otsu’s method to obtain an adaptive threshold $k^*$ by maximizing inter-class variance,
\[
k^* = \arg\max_{1 \le k < L} \sigma_B^2(k),
\]
and construct a background mask $M_{\text{bg}}$ by selecting pixels whose differences fall below $k^*$. We then compute metrics only on the masked background region. This provides a targeted measure of whether the model preserves non-edited context while performing the intended semantic edits.

\begin{figure*}
    \centering
    \includegraphics[width=1\linewidth]{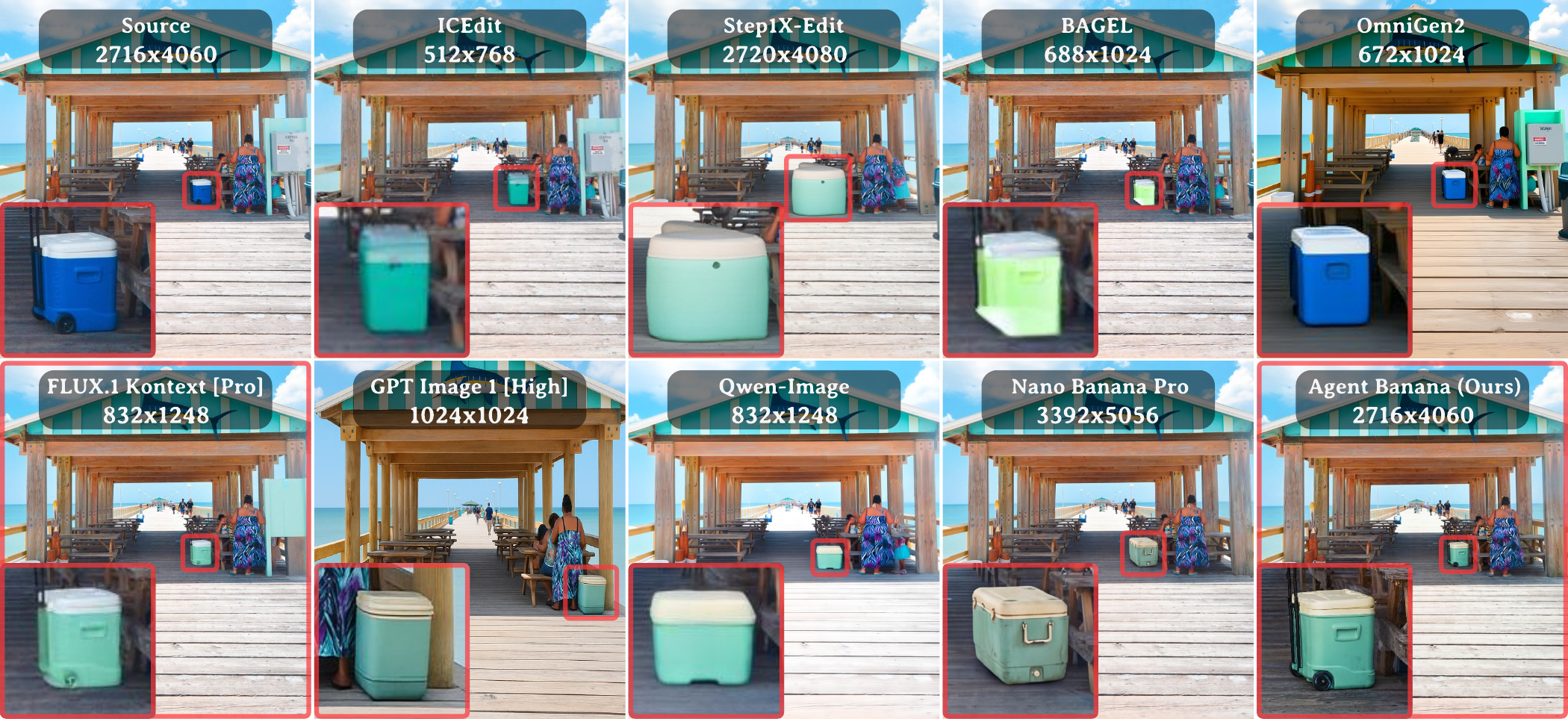}
    \vspace{-5mm}
    \caption{\textbf{Qualitative Comparison of Editing Fidelity.} We utilize the instruction \textit{"...And change that little bright blue cooler under the shelter to a softer sea‑foam green with a creamy top ..."} to guide the editing process. While the prompt solely targets color modification, baseline models exhibit significant limitations: they often suffer from reduced resolution, introduce unwanted structural changes (modifying shape or position), or fail to apply the target color change. By leveraging our agent's superior interpretation capabilities, our method accurately captures the instruction's focus while preserving the integrity of the original image.}
    \label{fig:hires}
\end{figure*}

\begin{table*}[!t]
    \centering
    \scriptsize
    \caption{\textbf{Quantitative Comparison of Image Editing Performance.} We evaluate models on HDD-Bench focusing on image fidelity (PSNR$_\text{OM}$, SSIM$_\text{OM}$, LPIPS$_\text{OM}$), instruction adherence (Instruct-Following, Image Consistency), and support for high-resolution (4K) editing. \textbf{Agent Banana} achieves state-of-the-art performance, balancing precise instruction execution with high visual fidelity, and is natively capable of processing at 4K resolution.}
    \resizebox{0.99\linewidth}{!}{
    \begin{tabular}{l|ccccc|cccc|c}
        \toprule
        \textbf{Model} & \multicolumn{5}{c|}{\textbf{HDD Bench}} & \multicolumn{4}{c|}{\textbf{ImgEdit}} & \textbf{4K} \\
        & \bf PSNR$_\text{OM}$  $\uparrow$& \bf SSIM$_\text{OM}$  $\uparrow$& \bf LPIPS$_\text{OM}$  $\downarrow$& \bf IF $\uparrow$ & \bf IC $\uparrow$ & \bf Add $\uparrow$ & \bf Adj. $\uparrow$ & \bf Repl. $\uparrow$ & \bf Rem. $\uparrow$ & \bf Res  \\
        \midrule
        ICEdit~\cite{zhang2025context} & \textbf{29.21} & \underline{0.80} & \underline{0.14} & 0.595 & 0.687 & 3.58 & 3.39 & 3.15 & 2.93  & -- \\
        Qwen-Image~\cite{wu2025qwen} & 23.62 & \underline{0.80} & \underline{0.14} & 0.845 & 0.807 & 4.38 & 4.16 & 4.66 & 4.14 & -- \\
        OmniGen2~\cite{wu2025omnigen2} & 23.59 & 0.72 & 0.23 & 0.545 &0.655 & 3.57 & 3.06 & 3.74 & 3.20  & -- \\
        BAGEL~\cite{deng2025bagel} & 26.93 & 0.79 & 0.17 & 0.676 & 0.723 & 3.56 & 3.31 & 3.3 & 2.62  & -- \\
        Step1X-Edit~\cite{liu2025step1x} & 25.82 & 0.77 & 0.19 & 0.808 & 0.797 & 3.88 & 3.14 & 3.40 & 2.41  &  \checkmark \\
        \midrule
        FLUX.1 Kontext [Pro]~\cite{labs2025kontext} & 25.98 & 0.74 & 0.17 & 0.845 & 0.702 & 4.25 & 4.15 & \underline{4.56} & 3.57 & -- \\
        GPT Image 1 [High]~\cite{gptimage} & 19.20 & 0.54 & 0.33 & \underline{0.882} & 0.727 & \textbf{4.61} & 4.33 & 4.35 & 3.66  & -- \\
        Nano Banana Pro~\cite{google2025gemini3imagepro} & 26.62 & 0.72 & \underline{0.14} & \textbf{0.911} &\underline{0.861} & \underline{4.58} & \underline{4.56} & 4.55 & \underline{4.39} & \checkmark \\
        \midrule
        \bf Agent Banana (ours) & \underline{28.40} & \textbf{0.84} & \textbf{0.12} & 0.849 & \textbf{0.871} & \underline{4.58} & \textbf{4.59} & \textbf{4.62} & \textbf{4.60} & \checkmark \\
        \bottomrule
    \end{tabular}
    }
    \label{tab:benchmark}
\end{table*}

\section{Experiments}
\subsection{Experimental Setup}
In our experiments, we employ \textbf{GPT-5-mini} as the foundational Large Language Model (LLM) powering both the Planner and Executor agents. To endow the agents with robust visual generation and editing capabilities, we construct a comprehensive toolset integrating state-of-the-art visual models, including both open-source and private models for high-quality generation and editing, complemented by \textbf{GPT-5-mini} for visual verification. To ensure a fair comparison, when Nano Banana Pro is used as the underlying image model, our gains reflect agentic scaffolding (decomposition, masking, verification) rather than changes to the generator weights; instead, we compare against the Nano Banana Pro and other baseline models operating without our multi-step workflow.

\subsection{Performance on Multi-turn Editing}
To comprehensively evaluate the performance of Agent Banana, we benchmark it against representative image editing models, including the closed-source commercial model \textbf{Flux.1 Kontext}~\cite{labs2025kontext}, \textbf{Nano Banana Pro}~\cite{google2025gemini3imagepro}, and \textbf{GPT-Image-1 [High]}~\cite{gptimage}. We adopt the standard metrics defined by \textbf{HDD-Bench}, covering editing accuracy (${s}_{\text{instruction following}}$), Otsu-Masked PSNR (${s}_{\text{$\text{PSNR}_\text{OM}$}}$), and the final composite score ($s_{\text{final}}$). Detailed quantitative comparisons are presented in Table~\ref{tab:benchmark}.

Given that this benchmark focuses on \textit{multi-turn sequential editing tasks}, we report the average score across all interaction turns as the final performance metric. Notably, our dataset consists entirely of \textbf{4K-resolution images}, posing a significant challenge to the high-resolution processing capabilities of the models. For baselines that downsample inputs during processing, we explicitly denote their maximum supported resolution in the results table and evaluate them after upsampling the output back to 4K.

The results indicate that Agent Banana not only achieves competitive scores against the baselines but, crucially, is one of only two models capable of maintaining high fidelity at \textbf{4K native resolution}. This validates the effectiveness of our proposed Image Layer Decomposition mechanism in preventing detail loss during high-resolution editing.

\subsection{Performance on Single-turn Editing}

In addition to evaluating long-horizon multi-turn capabilities, we assess the foundational performance of Agent Banana on single-turn editing tasks using the \textbf{ImgEdit-Bench}. This experiment aims to verify that our agent architecture, despite being designed for complex planning, maintains SOTA precision when handling atomic editing instructions.

We compare Agent Banana with mainstream single-turn editing models. As shown in Table~\ref{tab:benchmark}, our method achieves leading or comparable results across all metrics. This is primarily attributed to the Executor's precise control over tool parameters and the self-verification mechanism provided by the Quality Test modules.

\subsection{Ablation Study}

\textbf{Impact of Foundational LLM Capabilities.} 
We first investigate the sensitivity of system performance to the capabilities of the base model. Given that the base model directly dictates instruction understanding, task planning, and the accuracy of tool invocation, we experimented by replacing the kernels of the Planner and Executor with the smaller-scale \textbf{Qwen-3-8B}. Observations reveal that a weaker base model exhibits significant performance degradation when handling ambiguous instructions and long-sequence planning, frequently generating unparseable tool parameters or erroneous dependencies, leading to workflow interruptions. This confirms that robust reasoning capability is a prerequisite for agents handling complex multi-turn editing tasks.

\begin{figure}[t]
    \centering
    \includegraphics[width=0.8\linewidth]{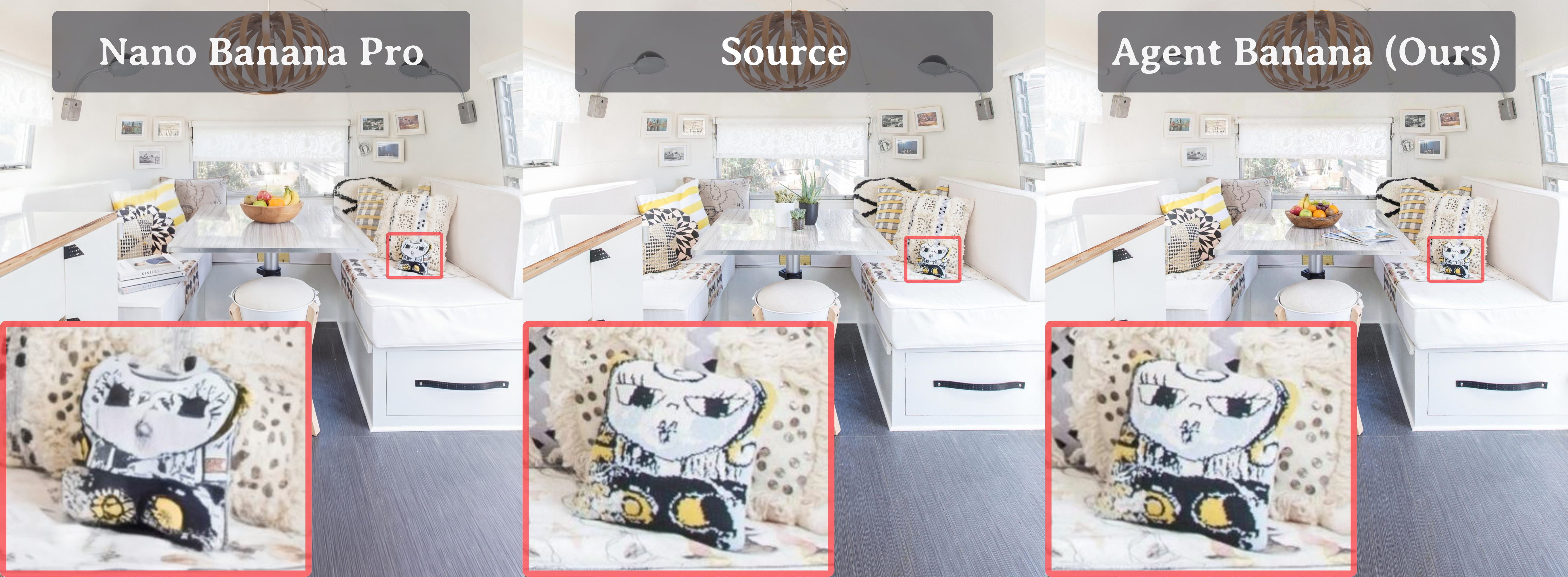}
    \caption{\textbf{Qualitative Comparison of Unedited Region Consistency.} Although the editing instruction does not target the sofa cushion, Nano Banana Pro distorts the original details due to global editing. In contrast, our method successfully maintains the visual consistency of the unedited regions.}
    \label{fig:QualCompUnchanged}
\end{figure}

\begin{figure}[t]
    \centering
    \includegraphics[width=1\linewidth]{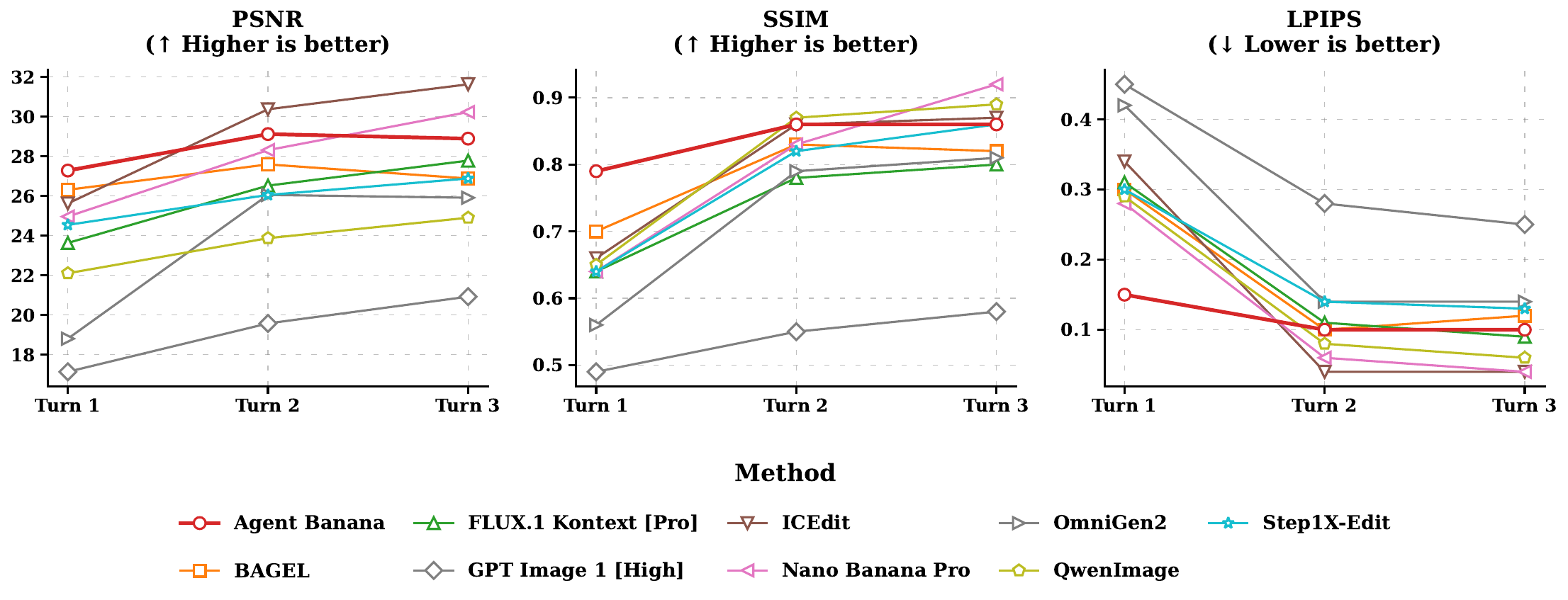}
        \vspace{-4mm}

    \caption{\textbf{Metric Comparison across Sequential Turns.} Agent Banana (red line) exhibits relatively better performance and consistent stability across all evaluated metrics. Compared to several other models, demonstrating its effectiveness in preserving image quality throughout the multi-turn process.}
    \label{fig:3}
\end{figure}

\subsection{Native-Resolution Editing Analysis}
A significant advantage of Agent Banana is its capability for \textbf{native-resolution editing}. Unlike existing baselines (e.g., FLUX.1 Kontext or Qwen-Image) that typically force inputs to be downsampled to 1k resolution, our method avoids this loss through a layered processing mechanism. As illustrated in Figure~\ref{fig:QualCompUnchanged}, for a high-resolution input of $2716 \times 4060$, baseline models lose substantial texture detail during the downsampling-upsampling process, whereas our method perfectly preserves the high-frequency information of the original image. However, baseline models exhibit significant limitations: they often suffer from reduced resolution, introduce unwanted structural changes (e.g., modifying object shape or position), or fail to apply the target color change. By leveraging our agent's superior interpretation capabilities, our method accurately captures the instruction's focus while preserving the integrity of the original image. This minimal-loss characteristic positions Agent Banana as a viable solution for professional-grade image editing tasks.

\subsection{On the Prior-Induced Editing Drift (PIED)}

We observe a subtle but important failure mode in multi-turn editing using generative editors: even when each turn appears highly realistic—sometimes indistinguishable to the eye—the purported “non-edited” regions (which are, in practice, repeatedly re-generated) can gradually drift toward the generator’s preferred texture and style statistics as turns accumulate.
We term this effect \textbf{Prior-Induced Editing Drift (PIED)}.
Figure~\ref{fig:3} shows that several baselines exhibit a steady \emph{increase} in $\text{PSNR}_\text{OM}$ on non-edited regions across turns, which can be misleading.
We hypothesize that PIED ``games'' this metric: repeated re-synthesis slightly re-renders the whole image, shrinking Otsu-partitioned background changes (thus inflating $\text{PSNR}_\text{OM}$) while faithfulness to the original input still degrades.
In contrast, \textbf{Agent Banana} keeps $\text{PSNR}_\text{OM}$ nearly constant across turns, matching qualitative observations of reduced accumulated artifacts and better preservation of high-frequency details and style in non-edited regions.
Overall, PIED suggests that \textbf{per-turn visual fidelity can decouple from long-horizon faithfulness}, and drift accumulation should be explicitly measured in evaluating multi-turn editors.

\section{Related Work}

\subsection{Instruction-based Image Editing}
Recent progress in instruction-based image editing has been driven by diffusion and autoregressive foundation models, such as GLIDE~\cite{nichol2021glide}, InstructPix2Pix~\cite{brooks2023instructpix2pix}, MagicBrush~\cite{zhang2023magicbrush}, Prompt-to-Prompt~\cite{hertz2022prompt}, and UltraEdit~\cite{zhao2024ultraedit}. Beyond these single-turn editors, emerging interactive systems (e.g., GPT-Image-1~\cite{openai2025imagegen} and Nano Banana~\cite{google2025gemini25flashimage}) indicate a shift toward multi-turn, context-aware interaction. To further strengthen fine-grained control, follow-up work explores attention manipulation~\cite{hertz2022prompt}, mask-based inpainting~\cite{avrahami2022blended}, and automatic region detection~\cite{couairon2022diffedit}; additionally, several methods decompose scenes into object-specific layers for more precise localized editing~\cite{monnier2021unsupervised, yang2025generative, wang2025layered}.

\subsection{Agentic Systems for Image Editing}

The exceptional reasoning and language capabilities of large language models (LLMs) have catalyzed rapid advances in agentic systems for interaction and task solving in complex environments. Paradigms exemplified by ReAct~\cite{yao2023react} establish a foundational framework by alternating reasoning and atomic actions within an iterative think–act loop. Meanwhile, Anthropic’s Model Context Protocol (MCP)~\cite{mcp_spec_2025_11_25} unifies the communication interface between LLMs and external tools, substantially improving the standardization and scalability of tool orchestration. Agentic perception–decision–action paradigms have long been explored in vision and learning via closed-loop or adaptive frameworks~\cite{rother2004grabcut,cubuk2019autoaugment,wang2021tent,zhu2023ghost,liu2025towards,chen2024restoreagent}, with VLMs increasingly serving as planners. For image/video restoration, AgenticIR and MoA-VR independently introduce VLM-integrated multi-agent repair paradigms~\cite{agenticir,liu2025moavrmixtureofagentsallinonevideo}. In creative photo retouching and task-oriented restoration, intelligent tool-invocation workflows such as JarvisIR, JarvisArt, 4KAgent, and JarvisEvo further demonstrate the effectiveness of agentic pipelines for restoration and editing~\cite{lin2025jarvisir,lin2025jarvisart,zuo20254kagent,lin2025jarvisevo}.

\section{Conclusion}

We introduce Agent Banana, a multi-agent, layer-aware framework for instruction-based image editing, together with HDD-Bench, a high-resolution multi-turn benchmark aligned with professional workflows. By coupling LLM planning, VLM perception, and layer-aware tool use, Agent Banana performs precise, rollback-safe edits on 4K images while preserving non-target regions, and consistently improves instruction following, edit locality, and multi-turn stability over strong non-agentic baselines. Beyond a single system and benchmark, we model editing as explicit state transitions on object-level graphs, enabling stepwise, verifiable evaluation and natural support for undo, branching, and long-horizon correction. Our scalable data engine further decouples state transitions from pixel rendering, making it practical to synthesize large-scale vision–language reasoning traces and edit histories. 

\paragraph{Impact Statement.}
This work advances instruction-based image editing toward professional workflows by emphasizing two properties that matter in real deployments: native high-resolution fidelity and multi-turn reliability. In particular, our benchmark and evaluation protocol provide stepwise, verifiable checks of what changed and what must remain invariant across turns, helping the community move beyond single-turn demos and toward diagnosing long-horizon failure modes such as over-editing, drift, and irreversible degradation.
At the same time, stronger editing capabilities can be misused to create misleading visual content or facilitate non-consensual manipulation.
 We therefore emphasize evaluation and auditing: our contributions are designed to measure controllability and detect failure accumulation rather than to optimize for unconstrained manipulation, and we encourage future systems built on this line of work to adopt provenance, consent, and disclosure mechanisms when applied to real-world media.

\clearpage
\bibliographystyle{plain}  
\small
\bibliography{main} 

\end{document}